\documentclass[10pt,twocolumn,letterpaper]{article}
\PassOptionsToPackage{table,xcdraw}{xcolor}
\usepackage{multirow}
\usepackage{svg}
\usepackage{mathtools}
\usepackage{bm}
\usepackage{amssymb}
\usepackage{capt-of}
\usepackage[pagenumbers]{iccv} % To force page numbers, e.g. for an arXiv version

\definecolor{iccvblue}{rgb}{0.21,0.49,0.74}
\usepackage[pagebackref,breaklinks,colorlinks,allcolors=iccvblue]{hyperref}

\newcommand{\nickname}{\textit{Epona}}

\def\paperID{5172} % *** Enter the Paper ID here
\def\confName{ICCV}
\def\confYear{2025}

\title{\textit{Epona}: Autoregressive Diffusion World Model for Autonomous Driving
}
\author{Kaiwen Zhang\textsuperscript{\normalfont 1,2*}\
\quad 
Zhenyu Tang\textsuperscript{\normalfont 1,3*} \
\quad
Xiaotao Hu\textsuperscript{\normalfont 1,5} \
\quad
Xingang Pan\textsuperscript{\normalfont 6} \\
\quad
Xiaoyang Guo\textsuperscript{\normalfont 1} \
\quad
Yuan Liu\textsuperscript{\normalfont 5} \
\quad
Jingwei Huang\textsuperscript{\normalfont 7} \
\quad
Li Yuan\textsuperscript{\normalfont 3} \
\quad
Qian Zhang\textsuperscript{\normalfont 1} \\
\quad
Xiao-Xiao Long\textsuperscript{\normalfont 4 \dag} \
\quad
Xun Cao\textsuperscript{\normalfont 4} \
\quad
Wei Yin\textsuperscript{\normalfont 1 \S}
\\
\vspace{-0.4cm}
\and
\textsuperscript{\normalfont 1}{\normalfont Horizon Robotics} \
\quad\textsuperscript{\normalfont 2}{\normalfont Tsinghua University} \
\quad\textsuperscript{\normalfont 3}{\normalfont Peking University} \\
\quad\textsuperscript{\normalfont 4}{\normalfont Nanjing University} \
\quad\textsuperscript{\normalfont 5}{\normalfont The Hong Kong University of Science and Technology} \\
\quad\textsuperscript{\normalfont 6}{\normalfont Nanyang Technological University} \
\quad\textsuperscript{\normalfont 7}{\normalfont Tencent Hunyuan} \ 
}
\begin{document}

\twocolumn[{
	\renewcommand\twocolumn[1][]{#1}
        \vspace{-10mm}
	\maketitle
	\vspace{-14mm}
	\begin{center}
        \includegraphics[width=\textwidth]{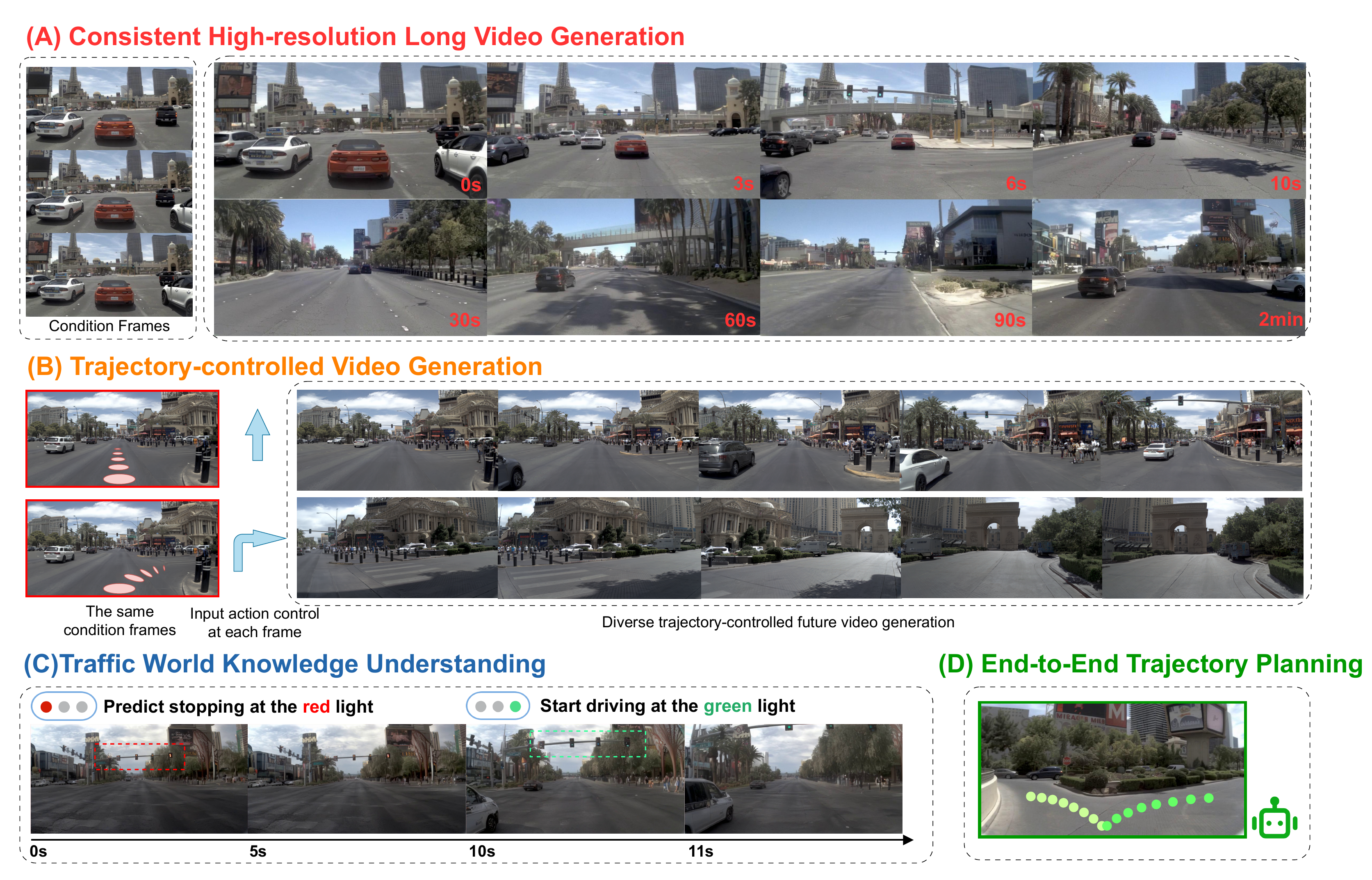}
	\end{center}
	\vspace{-7mm}
	\captionof{figure}{\textbf{Versatile capabilities of~\nickname.} Given historical driving context, our~\nickname~can generate consistent minutes-long future driving scenes at high resolution \textbf{(A)}. It can be controlled by diverse trajectories \textbf{(B)}, and understand real-world traffic knowledge \textbf{(C)}. In addition, our world model can predict future trajectories and serve as an end-to-end real-time motion planner \textbf{(D)}.}
	\label{fig:teaser}
	\vspace{3mm}
}]

% \twocolumn[{
%   \renewcommand\twocolumn[1][]{#1}
%   \maketitle
%   \vspace{-14mm}
%   \begin{center}
%     \includegraphics[width=\textwidth]{fig/teaser.pdf}
%     \vspace{-2mm}

%     % 手动添加 caption 和 label（非 \captionof）
%     \refstepcounter{figure}
%     {\small
%     \textbf{Figure~\thefigure.} \textbf{Versatile capabilities of~\nickname.}
%     Given historical driving context, our~\nickname~can generate consistent minutes-long future driving scenes at high resolution \textbf{(A)}.
%     It can be controlled by diverse trajectories \textbf{(B)}, and understand real-world traffic knowledge \textbf{(C)}.
%     In addition, our world model can predict future trajectories and serve as an end-to-end real-time motion planner \textbf{(D)}.}
%     \label{fig:teaser}
%   \end{center}
%   \vspace{3mm}
% }]

% \maketitle
\begin{abstract}

\footnote{
\textsuperscript{\normalfont *}These authors contributed equally to this work. 
\textsuperscript{\normalfont \dag}Project advisor. 
\textsuperscript{\normalfont \S}Project lead, Corresponding author.
}

Diffusion models have demonstrated exceptional visual quality in video generation, making them promising for autonomous driving world modeling. However, existing video diffusion-based world models struggle with flexible-length, long-horizon predictions and integrating trajectory planning.
This is because conventional video diffusion models rely on global joint distribution modeling of fixed-length frame sequences rather than sequentially constructing localized distributions at each timestep.
In this work, we propose \nickname, an autoregressive diffusion world model that enables localized spatiotemporal distribution modeling through two key innovations: 1) Decoupled spatiotemporal factorization that separates temporal dynamics modeling from fine-grained future world generation, and 2) Modular trajectory and video prediction that seamlessly integrate motion planning with visual modeling in an end-to-end framework.
Our architecture enables high-resolution, long-duration generation while introducing a novel chain-of-forward training strategy to address error accumulation in autoregressive loops.
Experimental results demonstrate state-of-the-art performance with 7.4\% FVD improvement and minutes longer prediction duration compared to prior works. The learned world model further serves as a real-time motion planner, outperforming strong end-to-end planners on NAVSIM benchmarks. Code will be publicly available at \href{https://github.com/Kevin-thu/Epona/}{https://github.com/Kevin-thu/Epona/}.

\end{abstract}
    
\section{Introduction}
\label{sec:intro}

Recently, with the rapid development of video generation models, world models have attracted significant attention and emerged as a powerful paradigm for physical world simulations and autonomous decision-making~\cite{ha2018wm, lecun2022wm, ding2024wmsurvey, agarwal2025cosmos}. 
These foundation models enable agents to understand inherent world knowledge and predict future dynamics, making them particularly promising for autonomous driving. 
Unlike traditional separate perception-planning pipelines, which require extensive annotations and explicit supervision, generative driving world models~\cite{hu2023gaia, gao2024vista, wang2023drivedreamer, chen2024drivinggpt, hu2024drivingworld, wang2023drivewm, zheng2024doe, gao2024magicdrivedit} integrate visual scene understanding with future prediction in a self-supervised manner, offering a new solution toward human-like, end-to-end autonomous driving.

Generative world models primarily fall into two categories: diffusion-based approaches and GPT-style autoregressive methods. The diffusion-based paradigm (e.g., Vista~\cite{gao2024vista}), while achieving impressive visual fidelity through joint distribution modeling of fixed-length videos~\cite{ddpm, song2021scorebased, ldm, svd}, fundamentally suffers from its inability to model per-timestep local distributions. This limitation manifests in critical deficiencies: failure to support variable-length long-range prediction crucial for dynamic world simulation, and infeasible trajectory planning due to the lack of mutlimodal prediction mechanism.

Conversely, GPT-style approaches~\cite{vaswani2017attention, gpt, gpt2, gpt3} address temporal flexibility through autoregressive next-token prediction (as seen in GAIA-1~\cite{hu2023gaia}).
However, the quantization and tokenization process significantly degrades visual quality and planning precision.
Moreover, the causal nature of autoregressive transformers constrains them to predicting only the next action rather than planning long-horizon trajectories~\cite{chen2024drivinggpt}, limiting their potential to serve as end-to-end driving planners.
Both paradigms reveal complementary shortcomings - diffusion models lack temporal decomposition while autoregressive transformers sacrifice continuous visual precision - highlighting the need for a unified framework that reconciles these divergent advantages for competent driving world modeling.

We introduce \nickname, an autoregressive diffusion world model that achieves high-resolution long-horizon video generation and accurate trajectory planning. Our core innovation comes from three key designs: 
1) \textbf{Decoupled spatiotemporal factorization.} 
While existing video diffusion methods model joint spatial-temporal distributions of past and future frames, we assume their temporal latent modeling lacks explicit causality constraints, leading to error accumulation in long sequences. \nickname~addresses this through spacetime-disentangled processing: A GPT-style transformer with causal attention handles temporal dynamics in compressed latent space, while twin diffusion transformers separately handle spatial rendering and trajectory generation.
2) \textbf{Asynchronous multi-modal generation}. 
Building upon this foundation, we decouple trajectory planning from visual generation through parallel denoising processes. Two specialized DiTs~\cite{peebles2023dit, flux2024} asynchronously generate
3-second vehicle trajectories and the single next future frame.
Both streams share flow-matching objectives~\cite{liu2022flow, lipman2023flowmatchinggenerativemodeling, albergo2023buildingnormalizingflowsstochastic} conditioned on the same temporal latent, ensuring alignment while preserving modality-specific optimizations.
3) \textbf{A chain-of-forward training strategy} addresses error accumulation and content drift in autoregressive loops. 

\nickname~offers several additional advantages.
1) \textbf{Long-horizon generation.} 
% With our proposed decoupled spatial-temporal modeling, next-frame autoregressive pipeline, and chain-of-forward training strategy,  
Our autoregressive diffusion model can achieve long-time generation over up to 2 minutes, significantly outperforming existing world models.
2) \textbf{Real-time trajectory planning.} 
The separate multi-modal generation architecture enables to perform trajectory planning solely while video prediction is deactivated, significantly reducing the inference FLOPS. This enables high-quality and even real-time trajectory planning, achieving rates of up to 20 Hz.
%First, with the factorized design, we can separate trajectory planning from visual generation. 
%Our tiny trajectory DiT is able to not only predict the next-frame action but also generate long trajectory plans in 0.05s, which allows our world model to seamlessly serve as a real-time motion planner.
\textbf{3) Visual detail preservation.} Our autoregressive formulation adopts continuous visual tokenizer instead of discrete ones, thus preserving rich scene details.

\begin{figure*}[t]
    \centering
	\includegraphics[width=\linewidth]{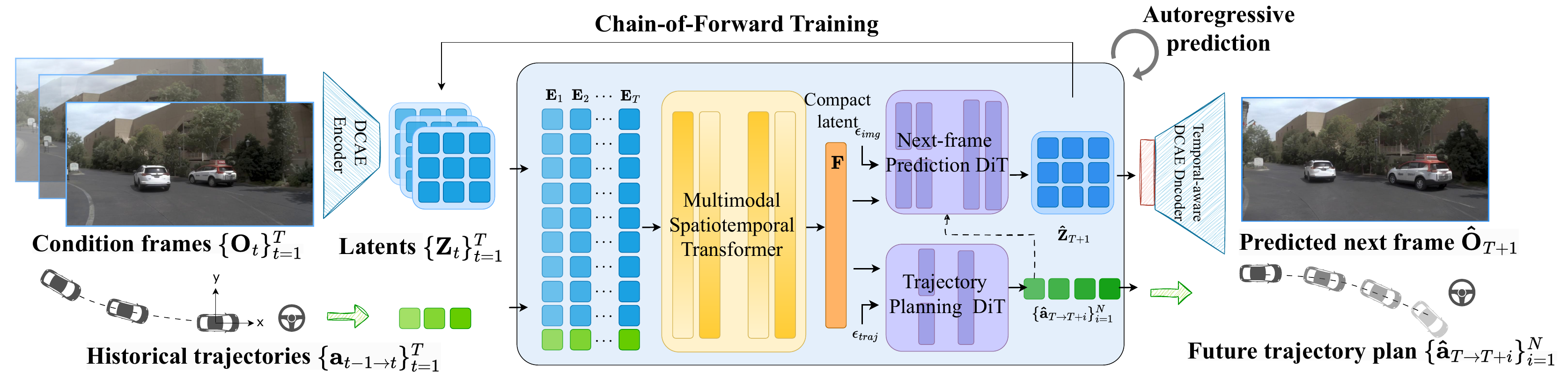}
        \vspace{-0.3cm}
	\caption{\textbf{Overview of \nickname.} Our world model utilizes a multimodal spatiotemporal transformer to process the historical context of the first $T$ frames and employs a next-frame prediction DiT to generate the frame at $T+1$ and a trajectory planning DiT to forecast the future $N$-frame pose trajectory. By adopting a chain-of-forward strategy, our approach enables high-quality and long-horizon video generation with an autoregressive manner.}
        \vspace{-0.3cm}
	\label{fig:pipeline}
\end{figure*}

Extensive experiments demonstrate the effectiveness and superiority of our world model. 
For video generation, our model achieves state-of-the-art FVD~\cite{fvd} on the NuScenes~\cite{caesar2020nuscenes} benchmark, surpassing the 
%best-performing driving world model~\cite{gao2024vista} by 7.4\%, 
best-performing Vista~\cite{gao2024vista} by 7.4\%,
while extending generation length from 15 seconds to over 2 minutes (600 frames).
Thanks to joint supervision with trajectory prediction,
\nickname~allows action control to simulate diverse driving scenarios, as shown in Fig.~\ref{fig:teaser} (B).
For motion planning, our method outperforms strong end-to-end planners on NAVSIM~\cite{Dauner2024navsim} without perception inputs (\eg, 3D boxes/lanes).
Notably, we observe that \nickname~learns essential traffic world knowledge (\eg, stop driving at red light) purely from self-supervised future prediction tasks, as shown in Fig.~\ref{fig:teaser} (C). 
This suggests that our world model can implicitly learn real-world driving dynamics, making it a promising direction for next-generation autonomous driving systems.

\section{Related Work}
\label{sec:relwork}
\subsection{World Models for Autonomous Driving}
Constructing real-world driving world models have drawn considerable attention in recent years, among which vision-centric approaches gain prominence due to their superior sensor flexibility, data accessibility, and more human-like representation forms. %especially those using visual modeling due to its better sensor flexibility, data accessibility, and richer and more human-like representation forms.
% Prior work was primarily based on a pre-trained diffusion model (\eg Stable Diffusion~\cite{ldm} or Stable Video Diffusion~\cite{svd}) and finetune on driving scenes.
Early efforts primarily focused on adapting pre-trained diffusion models (\eg, Stable Diffusion~\cite{ldm, svd}) to driving scenarios through fine-tuning.
However, these methods either lacked critical planning modules~\cite{gao2024vista, gao2024magicdrivedit, Gu2024DOMETD} or were limited to low-resolution, short-term generation~\cite{wang2023drivewm, wang2023drivedreamer, gao2023magicdrive, lu2023wovogen, yang2024generalized, Xing2025GoalFlowGF}, making them unsuitable for consistent long-range prediction and real-time planning.
% However, these diffusion-based methods either lack essential planning modules~\cite{gao2024vista, gao2024magicdrivedit} or restrict to low-resolution, short-term generation~\cite{wang2023drivewm, wang2023drivedreamer, gao2023magicdrive, lu2023wovogen, yang2024generalized}, hard to extend to consistent long-range prediction and real-time planning.
Recent work~\cite{hu2023gaia, chen2024drivinggpt, hu2024drivingworld, zheng2024doe} explored harnessing GPT-like architecture to unify visual and action modeling and achieved long-range autoregressive generation.
Yet, these methods require encoding images and trajectories into discrete tokens, which significantly degrades visual fidelity and trajectory precision.
Similarly, while the newly released Cosmos~\cite{agarwal2025cosmos} foundation model can serve as a driving world model, it does not introduce a new framework, facing the same limitations as the previous methods. 
In addition, its large parameter count and high computational demands limit its practicality.
In contrast, we propose a novel autoregressive diffusion world model framework for autonomous driving, enabling long-range autoregressive generation in continuous visual and trajectory representations.
% Similarly, while the newly released Cosmos~\cite{agarwal2025cosmos} foundation model can serve as a driving world model, it does not introduce new framework falling into the same issue as above and its large parameter count and heavy computational demands limit its practicality.
% However, these approaches require encoding images and trajectories into discrete tokens, leading to degraded visual fidelity and precision loss in trajectory planning.
% Recent released world foundation model Cosmos~\cite{agarwal2025cosmos} can also serve as driving world model, but Cosmos has large parameters and require heavy computational costs. 
% Besides, it does not introduce new world model framework, falling into the same issues as above.
% In contrast, we introduce a new autoregressive diffusion world model framework for autonomous driving,...

\subsection{Long Video Generation}
Long-term prediction is not only a key challenge for current video generation models but also a crucial capability for robust world models, as it reflects the model's ability to learn consistent environment dynamics and accurately simulate real-world temporal progression~\cite{ding2024wmsurvey}.
Since original video diffusion models (\eg SVD~\cite{svd}) are limited to fixed-length short clips generation, prior methods have explored extrapolating video length by noise rescheduling~\cite{qiu2023freenoise}, overlapped generation~\cite{wang2023genlvideo, Wang2024ZoLAZC} or hierarchical generation~\cite{yin2023nuwaxl}.
However, these techniques fail to resolve inherent model constraints, often resulting in inconsistencies and abrupt visual changes in long videos.
Autoregressive approaches~\cite{villegas2022phenaki, wang2024loong, chen2024drivinggpt} naturally support variable-length generation but suffer from quality degradation due to domain shift between teacher-forcing training and error accumulation in sampling.
GameNGen~\cite{gamengen} and DrivingWorld~\cite{hu2024drivingworld} introduce noise augmentation and random token dropout during training to alleviate the problem by simulating the error in sampling, but still limited to specific model architectures.
We propose a general chain-of-forward strategy allowing the model to directly learn inference errors during training, effectively reducing autoregressive drift.
Meanwhile, recent works such as Diffusion Forcing~\cite{chen2024diffusionforcing, song2025historyguidedvideodiffusion} and FIFO-Diffusion~\cite{kim2024fifo} explore integrating autoregressive generation in video diffusion by adjusting frame-wise noise levels and leveraging causal network designs. 
Our model adopts a similar causal temporal modeling strategy but redefines the architecture into a two-stage end-to-end framework, allowing joint generation of motion plans and next-frame images.
% Recently, Diffusion Forcing~\cite{chen2024diffusionforcing, song2025historyguidedvideodiffusion} and FIFO-Diffusion~\cite{kim2024fifo} also explore integrating autoregressive generation in diffusion models by using different frame-wise noise level and causal network design.
% Our model also employ the similar causal temporal modeling design choice to achieve variable-length generation but change the model architecture into a two-stage end-to-end pipeline.
\section{Method}
\label{sec:method}

In this section, we formally present the model framework and training techniques of~\nickname.
We begin with preliminaries on diffusion models in Sec.~\ref{subsec:prelim} and discuss our world model formulation design insights in Sec.~\ref{subsec:design}.
Then we introduce our proposed autoregressive diffusion world model framework in Sec.~\ref{subsec:wm}, including  three dedicated modules: a multimodal spatiotemporal transformer to capture historical context, a trajectory planning DiT to generate future 
3-seconds
trajectories, and a next-frame prediction DiT to generate the next-frame images.
To mitigate autoregressive drift and enable long-horizon video generation, we propose a simple yet effective chain-of-forward training strategy, detailed in Sec.\ref{subsec:cof}. 
Additionally, to enhance video quality, we introduce a temporal-aware DCAE decoder in Sec.\ref{subsec:dcae}. An overview of our method is illustrated in Fig.~\ref{fig:pipeline}.

\subsection{Preliminary}
\label{subsec:prelim}
% \noindent\textbf{Diffusion Models and Rectified Flow.}
Diffusion models~\cite{ddpm, song2021scorebased} are a family of powerful generative models that transform noise samples \( x_{(1)} \) drawn from a prior distribution \( p_1 = \mathcal{N}(\mathbf{0},\mathbf{I}) \) into data samples \( x_{(0)} \) from the target distribution in terms of a differentiable equation:
\begin{equation}
    dx_{(t)}=v_{\Theta}(x_{(t)}, t)dt, t\in [0,1],
\end{equation}
where velocity $v$ is parametrized by a neural network $\Theta$.
Rectified flow~\cite{liu2022flow, lipman2023flowmatchinggenerativemodeling, albergo2023buildingnormalizingflowsstochastic} proposes to define a straight probability path between $p_0$ and $p_1$ to improve training and sampling efficiency and optimize the network \( \Theta \) using a velocity prediction loss:
\begin{equation}
\label{eq:addn}
    x_{(t)}=(1-t)x_{(0)}+t\epsilon, \epsilon\sim\mathcal{N}(\mathbf{0}, \mathbf{I}),
\end{equation}
\begin{equation}
\label{eq:loss}
    \mathcal{L}_{RF} = \mathbb{E}_{x_{(0)}, \epsilon, t} \left[ \| v_\Theta(x_{(t)}, t) - ( x_{(0)} - \epsilon ) \|^2 \right],
\end{equation}
which has proven to be an effective and scalable approach in state-of-the-art image and video generation models~\cite{sd3, flux2024, kong2024hunyuanvideo}.
In~\nickname, we adopt the diffusion model and rectified flow objective for both next-frame image and trajectory generation.
Particularly, to enhance efficiency, we encode images into compact latents using a pre-trained deep compression encoder~\cite{dcae} and adopt a latent diffusion model~\cite{ldm} for image synthesis.

% \iffalse
\begin{figure}
	\includegraphics[width=\linewidth]{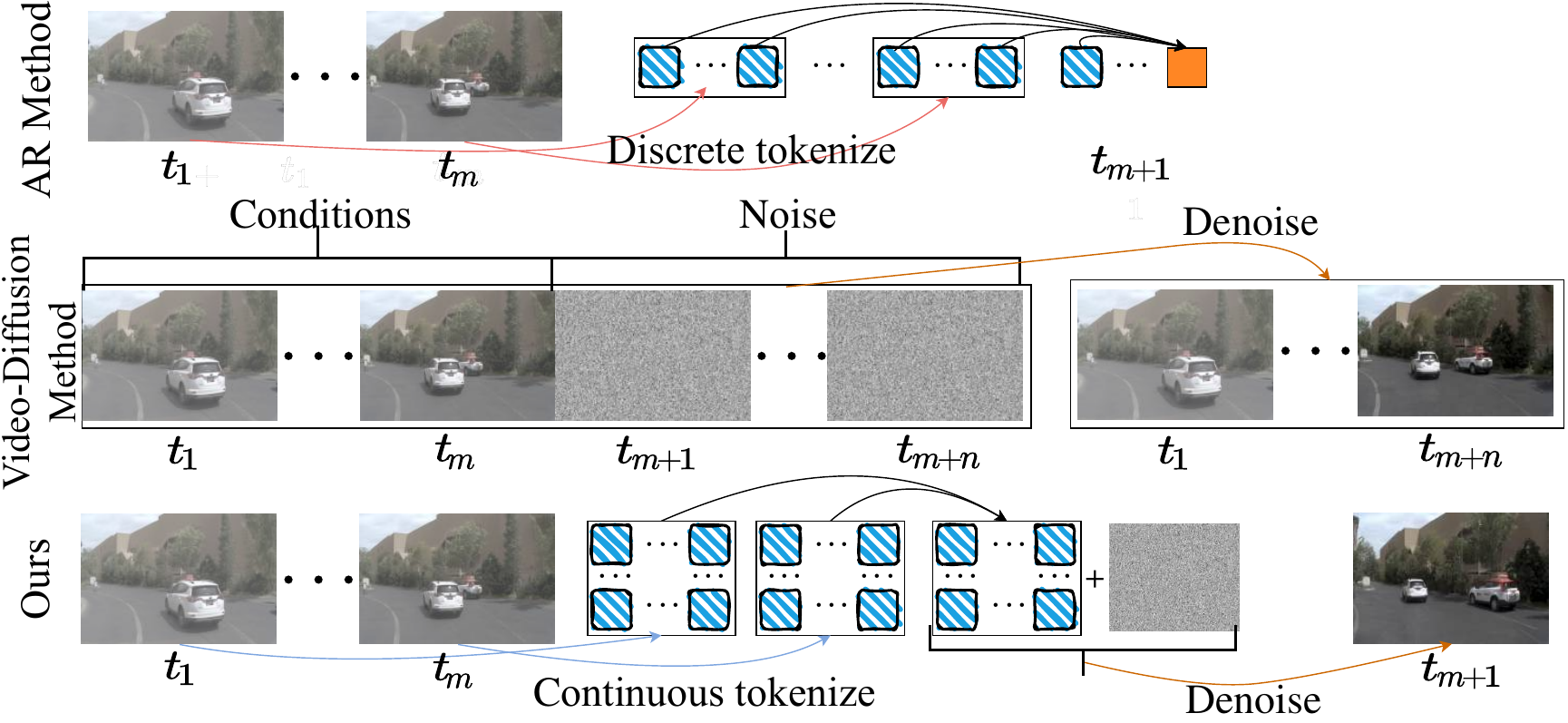}
        %\vspace{-0.3cm}
	\caption{\textbf{Comparison of Different World Modeling Formulation.} Up: Conventional autoregressive pipeline quantizes continous images into discrete tokens and perform next-token prediction iteratively. 
    Middle: The video-diffusion-based methods generate future $n$ frames simultaneously. 
    Down: Our method autoregressively predicts fine-grained future frames in continuous space.}
        %\vspace{-0.3cm}
	\label{fig: method cmp}
\end{figure}
% \fi

% \bigskip
% \noindent\textbf{Diffusion Transformers (DiT).}
% Recently, Diffusion Transformers (DiT)\cite{peebles2023dit} have emerged as a dominant backbone for diffusion models, replacing UNet\cite{Ronneberger2015UNetCN} due to their scalability and superior performance~\cite{sd3, flux2024, li2024hunyuandit, kong2024hunyuanvideo, chen2023pixartalphafasttrainingdiffusion}.
% In our model, we draw inspiration from the cutting-edge text-to-image model FLUX~\cite{flux2024} to design the backbone architecture for our second-stage diffusion transformers, enabling controllable, high-quality generation.

\subsection{Reformulation of World Model Designs}
\label{subsec:design}
In this section, we discuss different world model formulation design choices, as illustrated in Fig.~\ref{fig: method cmp}.
Given a sequence of previous front-view camera observations \(\{\mathbf{O}_t\}_{t=1}^{T}\) and the corresponding driving trajectory \(\{\mathbf{a}_{t-1\to t}\}_{t=1}^{T}\), the goal of driving world moels is to predict future driving dynamics based on historical context.
Here, each driving action \(\mathbf{a}_{t_1\to t_2} \coloneqq (\Delta\theta_{t_1\to t_2}, \Delta x_{t_1\to t_2}, \Delta y_{t_1\to t_2}) \in \mathbb{R}^{3}\) represents the vehicle's motion from \(t_1\) to \(t_2\), where \(\Delta\theta\) denotes the orientation change, and \((\Delta x, \Delta y)\) specify the relative displacement in the ego-coordinate frame. For consistency, we define \(\mathbf{a}_{0\to 1} = (0,0,0)\).
Existing methods tackle this problem by formulating world modeling in the following two ways:

\noindent\textbf{Video diffusion-based world models.}
Current leading driving world models, like Vista~\cite{gao2024vista}, formulates world modeling in the form of video diffusion models~\cite{svd}, which jointly capture the global spatiotemporal distribution of both past and a fixed-length future,
\[
p\left(\{\mathbf{O}_{T+i}\}_{i=1}^{n}, \{\mathbf{O}_t, \mathbf{a}_t\}_{t=1}^{T} \right).
\]
This formulation disrupts the causal temporal structure between historical observations and future predictions, limiting its ability to model real-world progressive dynamics and generate flexible-length long-term videos.

\noindent\textbf{GPT-based world models.}
Alternatively, autoregressive transformer-based world models~\cite{hu2023gaia, chen2024drivinggpt, hu2024drivingworld} discretize image observations into token sequences \(\mathbf{O}_{t}=[\mathbf{t}_1, \mathbf{t}_2,\cdots, \mathbf{t}_L]\) and model the conditional image distribution as a token-by-token prediction,  
\[
 \prod_{i=1}^{L} p(\mathbf{t}_i \mid \mathbf{t}_{<i}, \{\mathbf{O}_t, \mathbf{a}_t\}_{t=1}^{T}).
\]  
However, this independent token modeling weakens spatial correlations and the quantization process distortes high-frequency details, leading to degraded generation quality.  

\noindent\textbf{Our approach.} In contrast, we formulate world modeling as \textit{a sequential future prediction process in the temporal domain}.
Specifically, given past driving observations \(\{\mathbf{O}_t\}_{t=1}^{T}\) and the driving trajectory \(\{\mathbf{a}_{t-1\to t}\}_{t=1}^{T}\), our model predicts both a policy for future trajectory planning,  
\[
\pi\left(\{\mathbf{a}_{T\to T+i}\}_{i=1}^{n} \mid \{\mathbf{O}_t, \mathbf{a}_{t-1\to t}\}_{t=1}^{T} \right),
\]  
and a conditional distribution over the next-frame camera observation as a whole,  
\[
p\left(\mathbf{O}_{T+1} \mid \{\mathbf{O}_t, \mathbf{a}_{t-1\to t}\}_{t=1}^{T}, \mathbf{a}_{T\to T+1} \right).
\]  
The next-frame prediction is conditioned on either a model-predicted action or an externally provided action \(\mathbf{a}_{T\to T+1}\). 
By decoupling causal temporal modeling from fine-grained future prediction, our model can generate flexible-length long videos autoregressively in continuous representations.
Moreover, by factorizing trajectory planning from visual generation, our model can seamlessly serve as a real-time motion planner, bridging a critical gap between current driving world models and end-to-end motion planners.

\subsection{Epona: Autoregressive Diffusion World Model}
\label{subsec:wm}
% In~\nickname, we propose a factorized world model framework that decouples history causal modeling and fine-grained future world prediction, allowing for autoregressive next-frame generation and trajectory planning in continuous representations.

Based on the reformulated world modeling design, we propose~\nickname, an autoregressive diffusion world model for autonomous driving.
Our framework consists of three key components. First, a Multimodal Spatiotemporal Transformer (MST) encodes historical context \(\{\mathbf{O}_t, \mathbf{a}_t\}_{t=1}^{T}\) into a compact latent representation, effectively capturing environmental context and driving dynamics. 
Then, based on the historical latents, we employ two specialized diffusion transformers to predict fine-grained future details, including a tiny Trajectory Planning Transformer (TrajDiT) that models the policy \(\pi\) for trajectory planning, and a Next-frame Prediction Transformer (VisDiT) that models the visual distribution \(p\) for future image generation.
This modular design enables a range of autonomous driving applications. For instance, MST and VisDiT can be used independently for controllable driving simulations, while MST and TrajDiT facilitate real-time motion planning.

% \bigskip
\noindent\textbf{Multimodal Spatiotemporal Transformer (MST).}
Given the encoded past driving scenes \(\{\mathbf{Z}_t\}_{t=1}^{T}\) and trajectory \(\{\mathbf{a}_{t-1\to t}\}_{t=1}^{T}\), we introduce a multimodal spatiotemporal transformer to effectively integrate temporal dynamics and multimodal information from historical context for future prediction. 
Inspired by prior work in video generation and world modeling~\cite{Blattmann2023AlignYL, ma2024lattelatentdiffusiontransformer, hu2024drivingworld}, our approach employs interleaved multimodal spatial attention layers and causal temporal attention layers. 
This design progressively incorporates historical information into a compact latent representation while significantly reducing memory consumption compared to full-sequence attention. Additionally, this design naturally supports historical contexts of variable length.

Specifically, we first project the flattened visual latent patches \(\mathbf{Z} \in \mathbb{R}^{B\times T\times L\times C}\) and action sequences \(\mathbf{a} \in \mathbb{R}^{B\times T\times 3}\) into an embedding space. Then we concatenate them along the spatial dimension and add temporal positional embeddings to obtain the latent embedding sequence \(\mathbf{E} \in \mathbb{R}^{B\times T\times (L+3)\times D}\). This sequence is processed through interleaved multimodal spatiotemporal layers as follows (using \texttt{einops}~\cite{einops} notation):  
\begin{align*}
   \mathbf{E} &\leftarrow \texttt{rearrange}(\mathbf{E}, \texttt{(b\ t)\ l\ c\ $\to$ (b\ l)\ t\ c)}) \\
   \mathbf{E} &\leftarrow \texttt{CausalTemporalLayer}(\mathbf{E},\ \texttt{CausalMask}) \\
   \mathbf{E} &\leftarrow \texttt{rearrange}(\mathbf{E}, \texttt{(b\ l)\ t\ c\ $\to$ (b\ t)\ l\ c)}) \\
   \mathbf{E} &\leftarrow \texttt{MultimodalSpatialLayer}(\mathbf{E}),  
\end{align*}  
where \(B\) is the batch size, \(T\) is the number of conditioning frames, \(L = H \times W\) is the number of flattened latents in an image, \(C\) is the image latent channel dimension, \(D\) is the embedding dimension, and \texttt{CausalMask} is the triangular causal attention mask. 
Finally, we use the latent embedding of the last frame, $\mathbf{F} \in \mathbb{R}^{B \times (L+3) \times D}$, as the compact latent representation for the next-stage prediction. After training, this embedding encapsulates the historical context $\{\mathbf{O}_t, \mathbf{a}_{t-1\to t}\}_{t=1}^{T}$. %as formulated in Equations~\ref{eq:plan} and~\ref{eq:image}.

% \bigskip
\noindent\textbf{Trajectory Planning Diffusion Transformer (TrajDiT).}
TrajDiT predicts future trajectories using a tiny diffusion transformer. 
Following the DiT frameworks in most advanced open-source text-to-image and video generation models~\cite{flux2024, kong2024hunyuanvideo}, we adopt a Dual-Single-Stream architecture. 
In the dual-stream phase, the historical latent representation $\mathbf{F}$ and trajectory data are processed independently through transformer blocks, with only attention operations linking them. 
In the single-stream phase, they are concatenated to pass through subsequent transformer blocks for effective information fusion.
Detailed architecture can be found in the supplementary material.

During training, we add noise to the target trajectories $\overline{\mathbf{a}}\in \mathbb{R}^{B\times N\times 3}$ using Eq.~\ref{eq:addn}.
The model then predicts velocity $v_{traj}$ conditioned on $\mathbf{F}$, where $N$ is the planning horizon. We optimize using the rectified flow loss:
\begin{equation} \mathcal{L}_{traj} = \mathbb{E}_{\overline{\mathbf{a}}, \epsilon, t} \left[ \| v_{traj}(\overline{\mathbf{a}}_{(t)}, t) - (\overline{\mathbf{a}} - \epsilon) \|^2 \right]. \end{equation}
For inference, random Gaussian noise is iteratively denoised conditioned on $\mathbf{F}$ to generate future trajectory plans.

\begin{figure}[t!]
    \centering
	\vspace{-0.2cm}\includegraphics[width=0.9\linewidth]{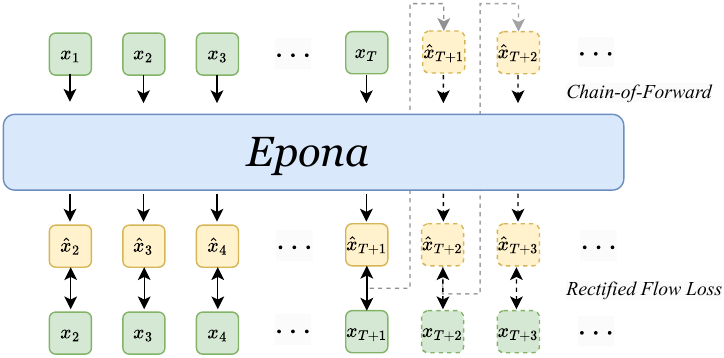}
        \vspace{-0.1cm}
	\caption{\textbf{Concept illustration of our training process.} Here $x$ can be either image latents or trajectories.}
        \vspace{-0.4cm}
	\label{fig:training}
\end{figure}

\noindent\textbf{Next-frame Prediction Diffusion Transformer (VisDiT).}
VisDiT has a similar architecture as TrajDiT, with an additional modulation~\cite{peebles2023dit} branch for action control $\mathbf{a}_{T\to T+1}$.
We also use the flow loss for visual supervision:
\begin{equation}
    \mathcal{L}_{vis} = \mathbb{E}_{\mathbf{Z}_{T+1}, \epsilon, t} \left[ \| v_{vis}(\mathbf{Z}_{T+1 (t)}, t) - ( \mathbf{Z}_{T+1} - \epsilon) \|^2 \right],
\end{equation}
Together, the total loss jointly optimizes the entire world model:
\begin{equation}
    \mathcal{L} = \mathcal{L}_{traj} + \mathcal{L}_{vis}.
\end{equation}
During inference, VisDiT denoises $\mathbf{\hat{Z}}_{T+1}$, conditioned on $\mathbf{F}$ and the action either predicted by TrajDiT or provided by user.
The latents are then decoded using the DCAE decoder to generate the next-frame image $\mathbf{\hat{O}}_{T+1}$.

\begin{table*}[ht]
\vspace{-1mm}
\centering
\caption{\textbf{Comparisons of generated videos on the NuScenes~\cite{caesar2020nuscenes} validation set.} 
Our model achieves state-of-the-art FVD score compared to existing driving world models, while extending the video length to over two minutes. *The max duration number indicates the horizon that produces plausible results, following existing methods.}
\vspace{-0.2cm}
\label{tab:fvd}
\resizebox{\textwidth}{!}{%
\begin{tabular}{lcccccccc}
\toprule
\textbf{Metric} 
 & DriveGAN~\cite{kim2021drivegan}
 & DriveDreamer~\cite{wang2023drivedreamer} & WoVoGen~\cite{lu2023wovogen} 
 & Drive-WM~\cite{wang2023drivewm} 
  & GenAD (OpenDV)~\cite{yang2024generalized} 
  & Vista~\cite{gao2024vista}  
  & DrivingWorld~\cite{hu2024drivingworld}
  & Ours \\
\midrule
\textbf{FID $\downarrow$} & 73.4  & 52.6 & 27.6 
& 15.8  & 15.4   & \textbf{6.9} & 7.4 & 7.5 \\
\textbf{FVD $\downarrow$} & 502.3
& 452.0 & 417.7 & 122.7  
& 184.0  & 89.4 & 90.9 & \textbf{82.8} \\
\textbf{Max Duration / Frames*} & N/A
& 4s / 48 & 2.5s / 5 & 8s / 16    
& 4s / 8  & 15s / 150 & 40s / 400 
& \textbf{120s / 600} \\
\bottomrule
\end{tabular}%
}
\end{table*}

\begin{figure*}[ht]
    \centering
	\includegraphics[width=\linewidth]{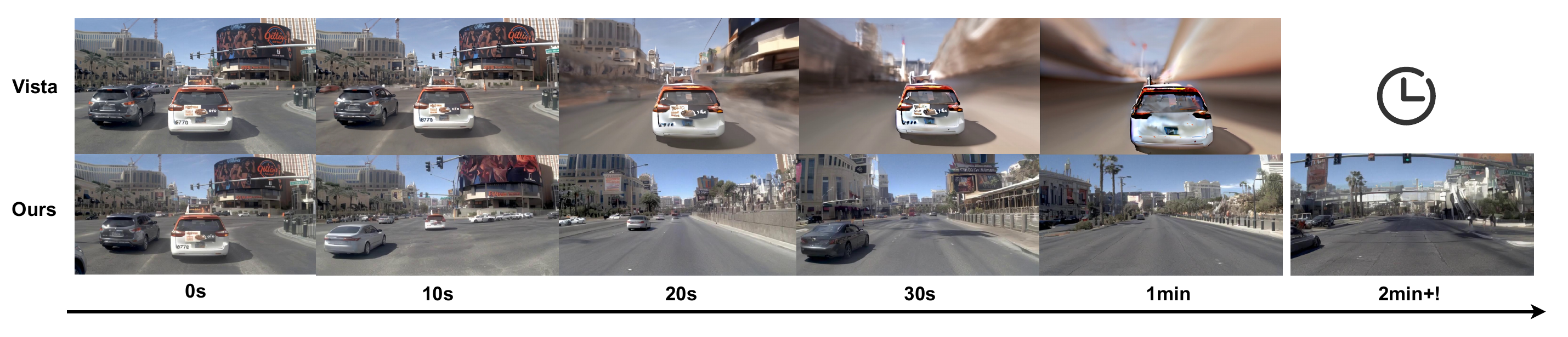}
        \vspace{-0.5cm}
	\caption{\textbf{Qualitative Comparison between Vista~\cite{gao2024vista} and~\nickname.} Zoom in for better views.}
        %\vspace{-0.1cm}
	\label{fig:qualitative}
\end{figure*}

\subsection{Chain-of-Forward Training}
\label{subsec:cof}

With our proposed autoregressive diffusion world model, we can autoregressively generate future videos frame by frame. However, long-term generation suffers from a long-standing autoregressive drift problem~\cite{gamengen}: 
during training, the model predicts the next frame using ground-truth historical context, whereas during inference, it relies on its own past predictions. 
This domain gap between teacher-forcing training and autoregressive sampling leads to error accumulation and rapid quality degradation.

To mitigate this, we introduce a chain-of-forward training strategy. 
Periodically, we perform multiple forward passes using self-predicted frames to enhance the model’s robustness to inference noise (see Fig.~\ref{fig:training}). 
Notably, to ensure training efficiency, instead of sampling next-frame latents from pure noise, we leverage the model-predicted velocity \( v_{\Theta} \) to estimate the denoised latents in one step: 
\begin{equation}
    \hat{x}_{(0)} = x_{(t)} + t v_\Theta(x_{(t)}, t)
\end{equation}
The estimated \( \hat{x}_{(0)} \), along with previous conditioned frames, is then used in the next forward pass to autoregressively generate subsequent frames. 
This process simulates prediction noise, helping the model adapt to deviations and improving long-term video generation quality.

\subsection{Temporal-aware DCAE Decoder}
\label{subsec:dcae}
Unlike conventional autoencoders that downsample images by a factor of 8, DCAE~\cite{dcae} progressively increases this to 32, reducing latent tokens by 16×. In our world model, we adopt DCAE for image encoding to improve training efficiency and reduce memory usage, enabling conditioning on longer historical contexts.

However, as an image autoencoder, DCAE lacks temporal interactions, causing flickering when decoding video frame by frame, which degrades visual quality. To address this, we propose a temporal-aware DCAE to enhance inter-frame consistency. Specifically, to maximize pretrained parameters while minimizing architectural changes, we introduce spatiotemporal self-attention layers before the DCAE decoder while keeping the encoder fixed during fine-tuning. This facilitates multi-frame interactions, greatly improving temporal consistency in generated videos.

% Unlike conventional mainstream autoencoders that typically compress the image size by a downsampling ratio of 8, DCAE~\cite{dcae} introduces a deep compression autoencoder that progressively increases the downsampling factor to 32. Compared to previous image autoencoders, DCAE greatly reduces the number of latent tokens by 16 times.  
% Therefore, we adopt DCAE to encode images into compact latents for our world model, which not only improves training efficiency but also significantly reduces memory consumption, allowing for conditioning on a longer historical context. 

% However, the original DCAE is an image autoencoder and lacks temporal interactions, leading to noticeable flickering when decoding the video frame by frame, which negatively impacts the visual quality of the world simulations. 
% To address this issue, we propose a temporal-aware DCAE to mitigate inter-frame flickering in generated videos. Specifically, to maximize the utilization of pretrained parameters and minimize architectural modifications, we only introduce multiple spatiotemporal self-attention layers before the DACE decoder and fix the encoder during fine-tuning. 
% This mechanism facilitates multi-frame information interactions, thus greatly improves the temporal consistency of the decoded videos.

\section{Experiment}
\label{sec:exp}
\subsection{Implementation Details}
\noindent\textbf{World Model. }
Our~\nickname~consists of 2.5  \textit{B} parameters, including a 12-layer multimodal spatiotemporal transformer with 1.3 \textit{B} parameters, a 12-layer next-frame prediction diffusion transformer with 1.2  \textit{B} parameters, and a 2-layer trajectory planning diffusion transformer with 50  \textit{M} parameters. 
It is trained on publicly available videos from the NuPlan dataset~\cite{caesar2021nuplan} and 700 scenes from the NuScenes dataset~\cite{caesar2020nuscenes} from scratch, all images are resized to 512×1024. 
We utilize the rectified flow~\cite{liu2022flow} objective for both video generation and trajectory planning tasks, training the entire model in an end-to-end manner.
The training was conducted on 48 NVIDIA A100 GPUs for nearly two weeks, with a total of 600k iterations and a batch size of 96. 
During training, we apply Chain-of-Forward every 10 steps, each time performing three forward passes.
We use the AdamW optimizer
with a learning rate of \(1 \times10^{-4}\) and
set the weight decay to \(5 \times 10^{-2}\).
For inference, we report our speed for each module on a single NVIDIA 4090 GPU in Table~\ref{tab:speed}. 
In all our experiments, we set our DiT sampling step to 100.
Notice that with MST and TrajDiT, our world model can seamlessly serve as a real-time motion planner.

\begin{table}[t]
% \vspace{-3mm}
\caption{\textbf{Inference speed.} We evaluate our inference speed for generating a 3-second trajectory and a \(512 \times 1024\) image per module on a single NVIDIA 4090 GPU.}
\centering
\resizebox{0.75\linewidth}{!}{
\begin{tabular}{lccc} 
\toprule
\multicolumn{1}{l}{DiT sampling steps}   &            
  \multicolumn{1}{l}{MST} & \multicolumn{1}{l}{TrajDiT} & \multicolumn{1}{l}{VisDiT} \\
\midrule
10  & $\sim$0.02s &  $\sim$0.03s & $\sim$0.3s \\
100 & $\sim$0.02s &  $\sim$0.3s & $\sim$2s \\
\bottomrule
\end{tabular}
}
\vspace{-0.3cm}
\label{tab:speed}
\end{table}

\begin{figure*}[t]
    \centering
	\includegraphics[width=0.95\linewidth]{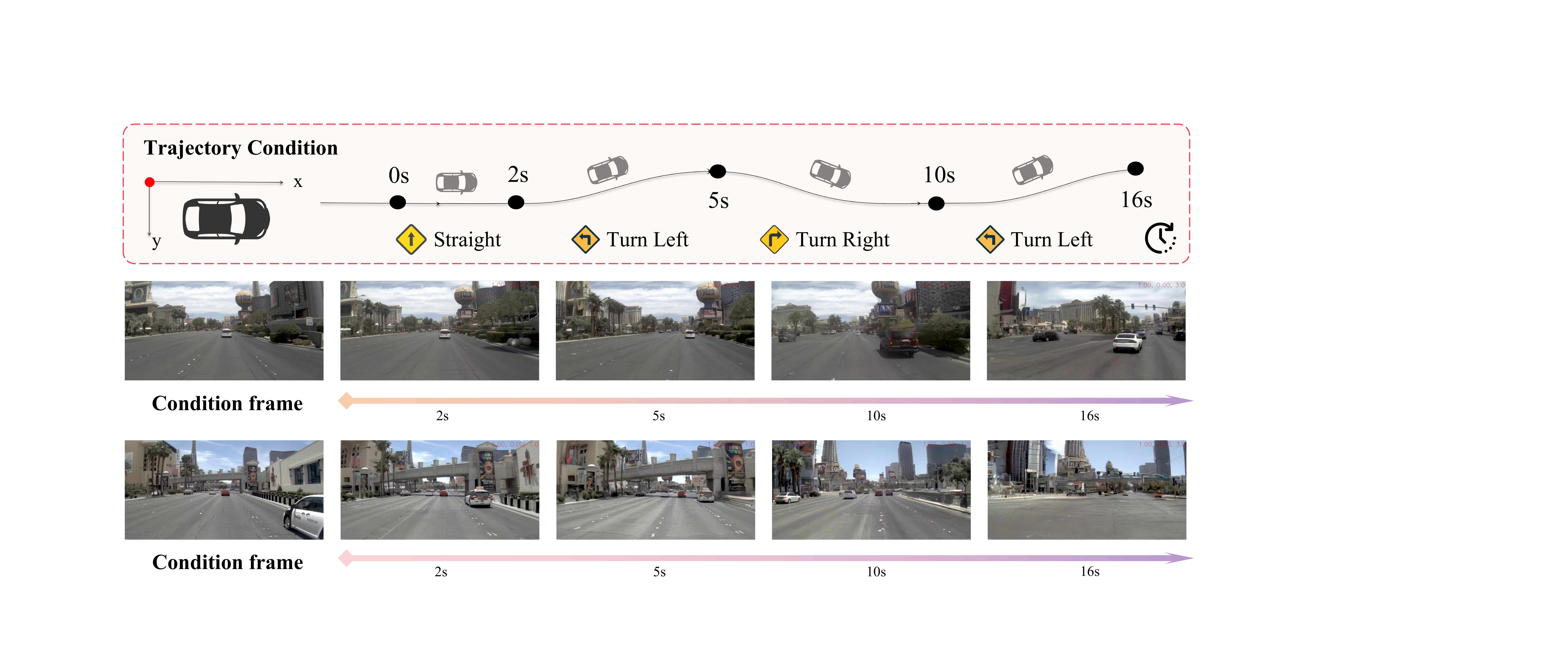}
        % \vspace{-0.3cm}
	\caption{\textbf{Trajectory-controlled video generation.} Our world model can generate controllable videos based on predefined trajectories. }
        \vspace{-0.3cm}
	\label{fig:control}
\end{figure*}

\noindent\textbf{Evaluations on video generation. }
We employ 1628 video clips from the NuPlan test set~\cite{caesar2021nuplan} and 1646 video clips from the NuScenes validation dataset~\cite{caesar2020nuscenes} for performance evaluation, respectively. During the test, our world model is conditioned on 10 consecutive past frames to generate the subsequent frame and repeat the process autoregressively to synthesize future video frames. We use the Frechet Video Distance (FVD)~\cite{fvd} and the Frechet Inception Distance (FID)~\cite{fid} to evaluate the quality of the generated videos.

\noindent\textbf{Evaluations on trajectory planning. }We evaluate trajectory planning using the NuScenes benchmark~\cite{caesar2020nuscenes} and the NAVSIM benchmark~\cite{Dauner2024navsim}. For the NuScenes, we use L2 error and collision rate as the evaluation metrics following the existing works~\cite{zheng2024doe, uniad, hu2022stp3} to evaluate the planning performance. L2 error
measures the L2 distance between the predicted and ground truth trajectories, while the collision rate measures the frequency of predicted trajectory intersections with objects. The NAVSIM benchmark assesses performance using the predictive driver model score (PDMS), derived from five factors, as shown in Table~\ref{tab:plan_nasim}. %: no at-fault collision (NC), drivable area compliance (DAC), time-to-collision (TTC), comfort (Comf.), and ego progress (EP). All metrics are computed based on a 4-second trajectory planning.

\begin{figure*}[ht]
    \centering
	\includegraphics[width=\linewidth]{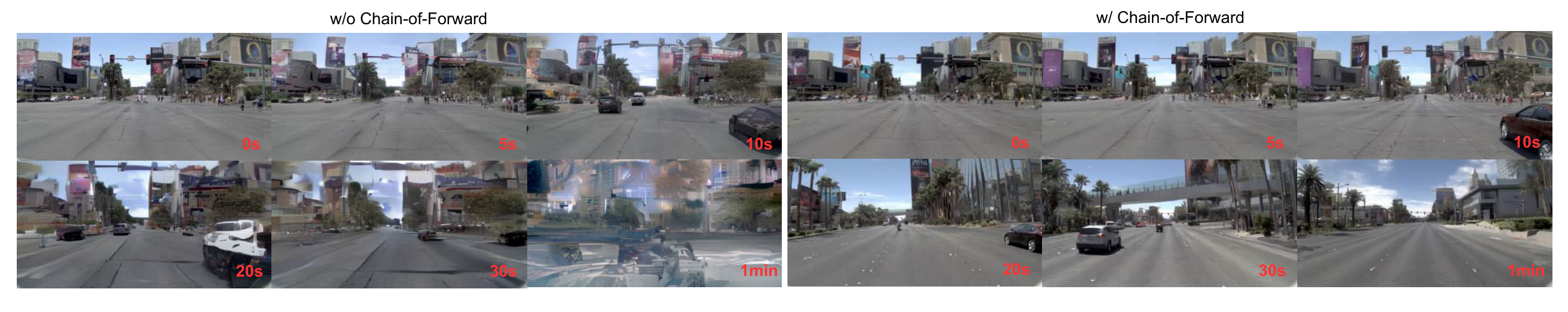}
        \vspace{-0.5cm}
	\caption{\textbf{Qualitative Comparison between long videos generated by models w/ and w/o Chain-of-Forward training.} Left: Visual quality deteriorates rapidly after 10–20 seconds. Right: The same driving scenes with Chain-of-Forward training maintain high visual quality, generating minute-long videos without significant degradation. Zoom in for better views.}
        %\vspace{-0.4cm}
	\label{fig:cof}
\end{figure*}

\begin{table*}[ht]
\setlength{\tabcolsep}{0.01\linewidth}
\caption{\textbf{End-to-end motion planning performance on the NuScenes~\cite{caesar2020nuscenes} dataset.} 
Note that our model achieves a low collision rate, demonstrating its understanding of basic traffic rules via simple next-frame prediction. $^*$ represents only using the front camera as input.}
\vspace{-3mm}
\centering
\resizebox{0.88\linewidth}{!}{
\begin{tabular}{l|lc|cccc|cccc}
\toprule
\multirow{2}{*}{Method} & \multirow{2}{*}{Input} & \multirow{2}{*}{Auxiliary Supervision} &
\multicolumn{4}{c|}{L2 (m) $\downarrow$} & 
\multicolumn{4}{c}{Collision Rate (\%) $\downarrow$} \\
&& & 1s & 2s & 3s & \cellcolor{gray!30}Avg. & 1s & 2s & 3s & \cellcolor{gray!30}Avg.  \\
\midrule
% IL~\cite{ratliff2006maximum} & LiDAR & None  & 0.44 & 1.15 & 2.47 & \cellcolor{gray!30}1.35 & 0.08 & 0.27 & 1.95 & \cellcolor{gray!30}0.77  \\
% NMP~\cite{zeng2019nmp} & LiDAR & Box \& Motion & 0.53 & 1.25 & 2.67 & \cellcolor{gray!30}1.48 & 0.04 & 0.12 & 0.87 & \cellcolor{gray!30}0.34  \\
% FF~\cite{hu2021safe} & LiDAR & Freespace  & 0.55 & 1.20 & 2.54 & \cellcolor{gray!30}1.43 & 0.06 & 0.17 & 1.07 & \cellcolor{gray!30}0.43  \\
% EO~\cite{khurana2022eo} & LiDAR & Freespace  & 0.67 & 1.36 & 2.78 & \cellcolor{gray!30}1.60 & 0.04 & 0.09 & 0.88 & \cellcolor{gray!30}0.33  \\
% \midrule
% \midrule
ST-P3~\cite{hu2022stp3} & Camera & Map \& Box \& Depth & 1.33 & 2.11 & 2.90 & \cellcolor{gray!30}2.11 & 0.23 & 0.62 & 1.27 & \cellcolor{gray!30}0.71  \\
UniAD~\cite{uniad} & Camera & { \footnotesize Map \& Box \& Motion \& Tracklets \& Occ}  & {0.48} & {0.96} & {1.65} & \cellcolor{gray!30}{1.03} & {0.05} & {\textbf{0.17}} & \textbf{0.71} & \cellcolor{gray!30}{\textbf{0.31}}  \\
OccNet~\cite{tong2023scene} & Camera & 3D-Occ \& Map \& Box & 1.29 & 2.13 & 2.99 & \cellcolor{gray!30}2.14 & 0.21 & 0.59 & 1.37 & \cellcolor{gray!30}0.72  \\
OccWorld~\cite{zheng2024occworld} & Camera & 3D-Occ & 0.52 & 1.27 & 2.41 & \cellcolor{gray!30}1.40 & 0.12 & 0.40 & 2.08 & \cellcolor{gray!30}0.87  \\
VAD-Tiny~\cite{vad}  & Camera & Map \& Box \& Motion  & 0.60 & 1.23 & 2.06 & \cellcolor{gray!30}1.30 & 0.31 & 0.53 & 1.33 & \cellcolor{gray!30}0.72  \\
VAD-Base~\cite{vad} & Camera & Map \& Box \& Motion & 0.54 & 1.15 & 1.98 & \cellcolor{gray!30}1.22 & 0.04 & 0.39 & 1.17 & \cellcolor{gray!30}0.53 \\
GenAD~\cite{zheng2024genad} & Camera & Map \& Box \& Motion & {\textbf{0.36}} & {\textbf{0.83}} & {\textbf{1.55}} & \cellcolor{gray!30}{\textbf{0.91}} & 0.06 & {0.23} & {1.00} & \cellcolor{gray!30}{0.43} \\
\midrule
Doe-1~\cite{zheng2024doe} & Camera$^*$ & QA & 0.50 & 1.18 & 2.11 & \cellcolor{gray!30}1.26 & 0.04 & 0.37 & 1.19 & \cellcolor{gray!30}0.53  \\
\textbf{\textbf{Ours}} & Camera$^*$ & None & 0.61 & 1.17  & 1.98 & \cellcolor{gray!30}1.25 & \textbf{0.01} &  0.22 & 0.85 & \cellcolor{gray!30}0.36  \\
\bottomrule
\end{tabular}%
}
\label{tab:plan_nusc}
\vspace{-3mm}
\end{table*}

\subsection{Evaluation of Video Generation}
\noindent\textbf{Quantitative Comparison of Generated Videos. }We present a quantitative comparison with existing methods on the NuScenes dataset~\cite{caesar2020nuscenes} in Table~\ref{tab:fvd}. Since most methods are not publicly available, we compare with the reported  results from their respective papers. Notably, the existing methods (e.g., Vista~\cite{gao2024vista}) fine-tune video diffusion models pre-trained on large-scale datasets, while our world model, including next-frame DiT, is trained from scratch. As shown in Table~\ref{tab:fvd}, our generated videos achieve state-of-the-art FVD scores, indicating smoother and more realistic video generation quality. Moreover, our world model can generate significantly longer video frames compared to existing approaches as shown in Table~\ref{tab:fvd}.

\noindent\textbf{Qualitative Comparison of Generated Videos. }We provide a qualitative comparison with the state-of-the-art open-source driving world model, Vista~\cite{gao2024vista}. Since Vista is a 25-frame fixed-length video diffusion model, we perform rollout to generate longer videos as illustrated in their paper. As shown in Fig.~\ref{fig:qualitative}, our~\nickname~generates consistent long-horizon driving scenes with high-fidelity visuals and detailed structures and vehicles.

\noindent\textbf{Trajectory-controlled Video Generation. }Fig.~\ref{fig:control} illustrates the pose controllability of our model. Given the predefined pose trajectory, the different condition frames can generate future frames that conform to the corresponding motion path, which is crucial for obtaining autonomous driving videos under extreme scenarios.

\noindent\textbf{Extra Long-range Video Generation. }\nickname~ combines the strengths of autoregressive and diffusion models, facilitating the generation of high-quality, long-duration videos conditioned on input frames. 
As shown in Fig.\ref{fig:qualitative} and Fig.\ref{fig:cof}, our model can autoregressively generate minute-long driving videos with high fidelity and consistency, without noticeable drift.
 %, continuously predicting new scenes without repeating past content, a common issue in SVD fine-tuning methods. 
% Moreover, our next-frame prediction DiT, ensures high fidelity with 3D consistency.
More long-term generation videos are provided in the supplementary materials.

\begin{table*}[ht]
\vspace{-1mm}
    \centering
    \caption{\textbf{End-to-end motion planning performance on the NAVSIM~\cite{Dauner2024navsim} test set.} NC: no at-fault collision. DAC: drivable area compliance. TTC: time-to-collision. Comf.: comfort. EP: ego progress. PDMS: the predictive driver model score. LAW\cite{li2024enhancing} is in the perception-free setting. Our world model outperforms strong end-to-end planners in the overall PDMS score.}
    \resizebox{0.8\textwidth}{!}{
    \begin{tabular}{lc|cc|ccc|c}
        \toprule
        \textbf{Method} & \textbf{Input} & \textbf{NC} $\uparrow$ & \textbf{DAC} $\uparrow$ & \textbf{TTC} $\uparrow$ & \textbf{Comf.} $\uparrow$ & \textbf{EP} $\uparrow$ & \textbf{PDMS} $\uparrow$ \\
        \midrule
        Human & / & 100 & 100 & 100 & 99.9 & 87.5 & 94.8 \\
        % Constant Velocity & / & 69.9 & 58.8 & 49.3 & 100 & 49.3 & 21.6 \\
        % Ego Status MLP & / & 93.0 & 77.3 & 83.6 & 100 & 62.8 & 65.6 \\
        \midrule
        UniAD\cite{uniad} & Camera & 97.8 & 91.9 & 92.9 & \textbf{100} & 78.8 & 83.4 \\
        PARA-Drive\cite{Weng2024paradrive} & Camera & \underline{97.9} & 92.4 & 93.0 & 99.8 & 79.3 & 84.6 \\
        LAW\cite{li2024enhancing} & Camera &96.4 & \textbf{95.4} & 88.7 & \underline{99.9} & \textbf{81.7} & 84.6 \\
        TransFuser\cite{Prakash2021transfuser} & Camera \& Lidar & 97.7 & 92.8 & 92.8 & \textbf{100} & 79.2 & 84.0 \\
        DRAMA\cite{yuan2024drama} & Camera \& Lidar & \textbf{98.0} & 93.1 & \textbf{94.8} & \textbf{100} & 80.1 & \underline{85.5} \\
        VADv2\cite{chen2024vadv2} & Camera \& Lidar & 97.2 & 89.1 & 91.9 & \textbf{100} & 76.0 & 80.9 \\
        \midrule
        Ours & Camera & \underline{97.9} & \underline{95.1} & \underline{93.8} & \underline{99.9} & \underline{80.4} & \textbf{86.2} \\
        \bottomrule
    \end{tabular}
    }
    \label{tab:plan_nasim}
    \vspace{-2mm}
\end{table*}

\subsection{Evaluation of Trajectory Planning}
For the NuScenes benchmark~\cite{caesar2020nuscenes}, we compare our world model with several existing methods, as shown in Table~\ref{tab:plan_nusc}. Although our model does not achieve the best results, it attains competitive performance without any additional supervision. It is worth noting that incorporating more supervision typically leads to better performance but cost expensive annotations. Additionally, similar to Doe-1~\cite{zheng2024doe}, our model only utilizes the front camera, whereas other methods rely on multi-view inputs for planning. As shown in Table~\ref{tab:plan_nusc}, our approach can generate reasonable trajectories while achieving the lowest collision rate for a 1-second horizon, which is crucial for long-term realistic video predictions. For the more challenging NAVSIM benchmark~\cite{Dauner2024navsim}, as shown in Table~\ref{tab:plan_nasim}, our method achieves state-of-the-art results in overall PDMS when conditioned on the past 2 seconds of observations to predict 4-second future trajectories, showing the strong motion planning capability.

\subsection{Ablation Study}
\noindent\textbf{Effect of Shared Latent for Multi-modal Joint Prediction.}
To assess the benefit of jointly modeling scene and trajectory via a shared latent representation, we conduct an ablation by disabling video prediction and training the model solely for trajectory prediction. 
This variant is evaluated on the NAVISIM test set.
As shown in Tab.~\ref{tab:cmp_navsim}, removing video prediction leads to a noticeable drop in planning performance. This result highlights the advantage of the shared latent $\mathbf{F}$, which encourages the world model to better capture complex driving dynamics by leveraging visual signals. 
This ablation confirms that coupling video and trajectory prediction within a unified latent space in world models significantly benefits downstream planning tasks.

\begin{table}[h!]
% \vspace{-4mm}
    \centering
    \caption{\textbf{Comparison of planning results on the NAVSIM test set.} Jointly predicting the next scene using a shared latent significantly improves planning performance.}
    % \vspace{-2.5mm}
    \resizebox{\linewidth}{!}{
    \begin{tabular}{l|cc|ccc|c}
        \toprule
        \textbf{Method} & \textbf{NC} $\uparrow$ & \textbf{DAC} $\uparrow$ & \textbf{TTC} $\uparrow$ & \textbf{Comf.} $\uparrow$ & \textbf{EP} $\uparrow$ & \textbf{PDMS} $\uparrow$ \\
        \midrule
        % DrivingGPT & \textbf{98.9} & 90.7 & \textbf{94.9} & 95.6 & 79.7 & 82.4 \\
        % \midrule
        Ours w/o Joint Training & 94.5 & 89.7 & 88.1 & \textbf{99.9} & 74.7 & 78.1 \\
        Ours & \textbf{97.9} & \textbf{95.1} & \textbf{93.8} & \textbf{99.9} & \textbf{80.4} & \textbf{86.2} \\
        \bottomrule
    \end{tabular}
    }
    \label{tab:cmp_navsim}
    % \vspace{-4mm}
\end{table}

\begin{figure}[t]
	\centering
        \vspace{-0.3cm}
	\includegraphics[width=\linewidth]{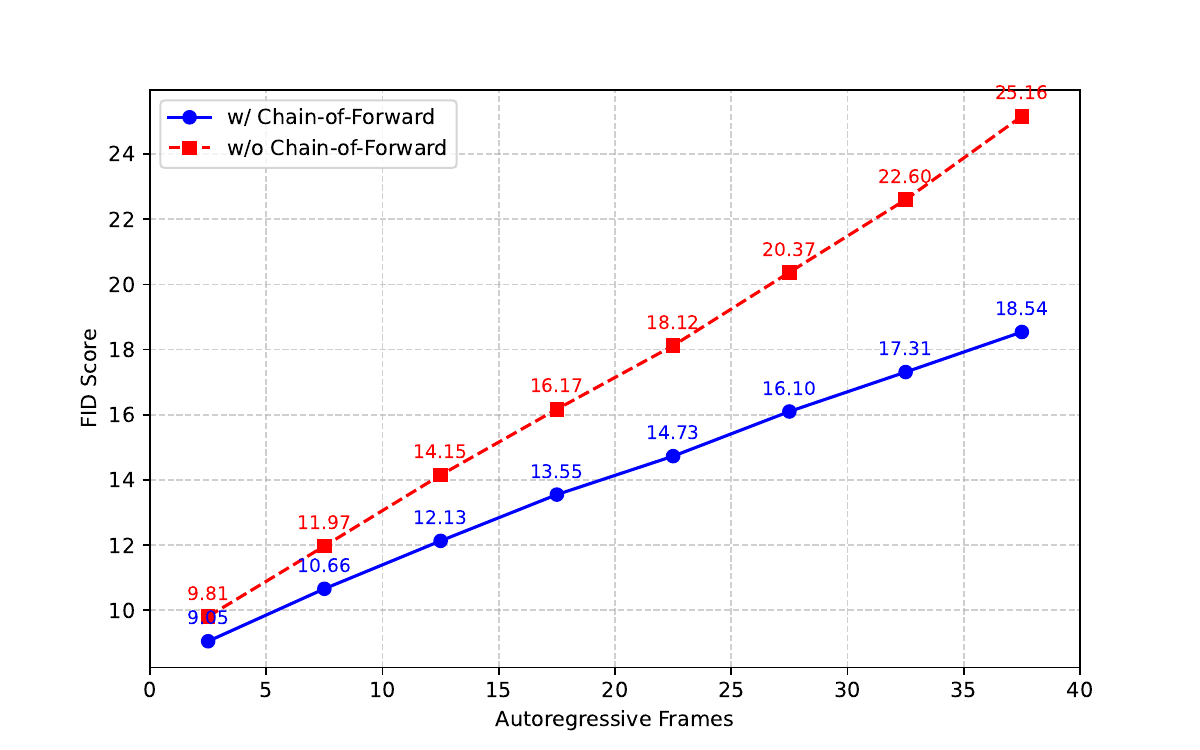}
        \vspace{-0.4cm}
	\caption{\textbf{Effect of Chain-of-Forward training.} FID comparison in NuPlan test set between models w/ and w/o Chain-of-Forward training strategy.}
	\label{fig:cof_fid}
        \vspace{-0.5cm}
\end{figure}

\noindent\textbf{Effect of Chain-of-Forward Training.} To evaluate the impact of our chain-of-forward strategy on model performance, we conduct an ablation study comparing results with and without this strategy. Given that our model iteratively generates the next frame, the chain-of-forward approach simulates potential inference errors during training, thereby enhancing the model's robustness. 
As shown in Fig.~\ref{fig:cof} and Fig.~\ref{fig:cof_fid}, as the model autoregressively generates longer sequences, the gap in visual quality and FID score between models with and without the chain-of-forward strategy becomes increasingly significant, validating its effectiveness in long-term video generation.

\noindent\textbf{Effect of Temporal-aware DCAE Decoder.} Considering that the original DCAE is an image-based autoencoder without temporal modeling capability, we incorporate a temporal interaction module before the DCAE decoder. As shown in Table~\ref{tab:dcae}, our world model achieves improved performance with the temporal module, effectively reducing flickering and enhancing the smoothness of the generated videos.

% \begin{table}[h!]
% % \vspace{-0.3cm}
% \caption{\textbf{Comparison of the Generated Videos \textit{w/} and \textit{w/o} Temporal-aware DCAE Decoder on NuScenes validation set.} Temporal-aware DCAE Decoder can mitigate flickering artifacts and improve smoothness in generated videos.}
% \centering
% % \vspace{-0.3cm}
% \resizebox{0.8\linewidth}{!}{
% \begin{tabular}{lccc} 
% \toprule
% \multicolumn{1}{l}{Method}   &            
%   \multicolumn{1}{l}{FVD$_{10}$ ↓} & \multicolumn{1}{l}{FVD$_{25}$ ↓} & \multicolumn{1}{l}{FVD$_{40}$ ↓} \\
% \midrule
% \textit{w/o} Temporal Module  & 58.11 & 89.93 & 114.34 \\
% Ours   & \textbf{48.67} & \textbf{82.83} & \textbf{106.43} \\
% \bottomrule
% \end{tabular}
% }
% % \vspace{-0.35cm}
% \label{tab:dcae}
% \end{table}

\begin{table}[ht]
\caption{\textbf{Comparison of the Generated Videos \textit{w/} and \textit{w/o} Temporal-aware DCAE Decoder Module on NuPlan~\cite{caesar2021nuplan} test set.} Temporal-aware DCAE Decoder can mitigate flickering artifacts and improve smoothness in generated videos.}
\centering
\resizebox{0.82\linewidth}{!}{
\begin{tabular}{lccc} 
\toprule
\multicolumn{1}{l}{Methods}   &            
  \multicolumn{1}{l}{FVD$_{10}$ ↓} & \multicolumn{1}{l}{FVD$_{25}$ ↓} & \multicolumn{1}{l}{FVD$_{40}$ ↓} \\
\midrule
\textit{w/o} Temporal Module  & 52.95 &  76.46 & 100.11 \\
Ours   & \textbf{50.77} &  \textbf{61.46} & \textbf{74.88} \\
\bottomrule
\end{tabular}
}
% \vspace{-0.3cm}
\label{tab:dcae}
\end{table}

\noindent\textbf{Effect of Different Context Length.} We gradually increase the length of conditioned frames to investigate its impact on model performance. As shown in Table~\ref{tab:conditon}, as the number of conditioned frames increases, our world model improves in FVD performance due to longer historical information. However, longer conditioned frames require handling extended sequences, which poses computational challenges. Given our model setting, 10 frames represent the upper limit for conditioning. Therefore, we ultimately select 10 frames as the conditioning length in our approach.

\begin{table}[ht]
\caption{\textbf{Comparison of different condition frames on NuPlan ~\cite{caesar2021nuplan} test set.}~\nickname~generates better videos when conditioning more frames.}
\centering
\resizebox{0.78\linewidth}{!}{
\begin{tabular}{lccc} 
\toprule
\multicolumn{1}{l}{Frame number}   &            
  \multicolumn{1}{l}{FVD$_{10}$ ↓} & \multicolumn{1}{l}{FVD$_{25}$ ↓} & \multicolumn{1}{l}{FVD$_{40}$ ↓} \\
\midrule
2  & 59.85 &  81.58 & 103.70 \\
5  & 55.46 &  71.28 & 86.76 \\
10   & \textbf{50.77} &  \textbf{61.46} & \textbf{74.88} \\
\bottomrule
\end{tabular}
}
\vspace{-0.4cm}
\label{tab:conditon}
\end{table}

% \noindent\textbf{\discuss{Effects of Multi-stage Training}}

\section{Conclusion}
\label{sec:concls}
We have presented~\nickname, an autoregressive diffusion world model for autonomous driving that jointly predicts high-fidelity future trajectories and driving scenes based on historical driving context. 
Thanks to our proposed decoupled spatiotemporal modeling and asynchronous multi-modal generation strategies, our model achieves high-quality and long-term prediction.
In addition, our model could serves as a real-time motion planner via performing trajectory planning.
We have demonstrated that our approach significantly advances the state of the art in driving world models, uncovering the large potential for building next-generation autonomous driving systems.

% Unlike that the traditional diffusion models construct global joint distribution of fixed-length frame sequences, we propose to disentangle temporal and spatial coherence in world modeling.

\newpage
{
    \small
    \bibliographystyle{ieeenat_fullname}
    \bibliography{main}
}

\clearpage
\appendix
\setcounter{page}{1}
\maketitlesupplementary
\begin{figure*}[!t]
    \centering
	\includegraphics[width=\linewidth]{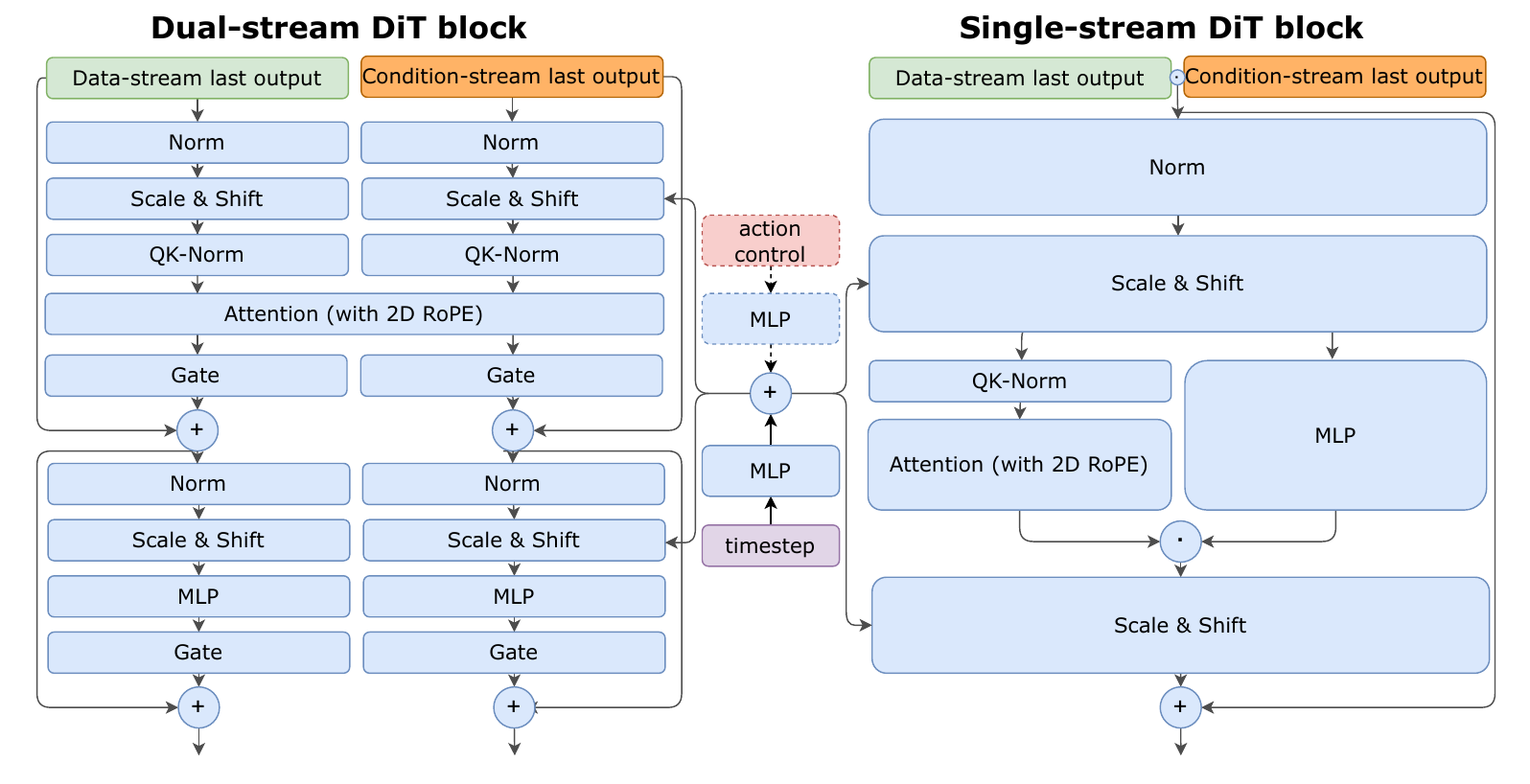}
        % \vspace{-0.3cm}
	\caption{\textbf{Detailed architecture of dual-stram DiT and single stream DiT blocks.} We use nearly identical architecures for both TrajDiT and VisDiT, modified from text-image and video DiT architecture from~\cite{flux2024, kong2024hunyuanvideo}.  Action control is only for VisDiT. }
        \vspace{-0.5cm}
	\label{fig:dit}
\end{figure*}

\section{Detailed Architecture of Dual-Single-Stream DiT}
We are inspired by recent state-of-the-art image and video generation architectures~\cite{flux2024, kong2024hunyuanvideo} and integrate dual-stream DiT blocks and single-stream DiT blocks to construct our TrajDiT and VisDiT. In the dual-stream DiT, condition information and noise are processed separately and interact only within the attention mechanism. In contrast, the single-stream DiT concatenates condition information and noise from the beginning for unified processing. Additionally, action control is mapped as an auxiliary control to obtain scale and shift parameters for adaptive modulation. The detailed architecture is illustrated in Fig.~\ref{fig:dit}.

\section{More Discussions with Related Works}

\paragraph{Comparison with GAIA-1~\cite{hu2023gaia}, DrivingGPT~\cite{chen2024drivinggpt}, and ADriver-I~\cite{adriver}.}
Compared to these multi-modal driving world models, our method adopts a fundamentally different architecture by directly integrating trajectory prediction into the video generation process via diffusion models. To the best of our knowledge, we are the \textit{first} driving world model to use diffusion models for generating continuous, multi-step action trajectories, which brings two key advantages:

\begin{enumerate}
    \item \textit{Multi-step vs. single-step prediction.} Unlike prior approaches that interleave single-step image and action generation using transformers, our model predicts an entire $N$-step future trajectory in one shot. This is particularly beneficial for real-time motion planning in autonomous driving.
    \item \textit{Continuous vs. discrete action representation.} While existing methods discretize continuous action spaces into tokens, our diffusion model generates high-resolution continuous trajectories directly, enabling more precise planning and control.
\end{enumerate}

Among these methods, only DrivingGPT reports NAVISIM planning metrics. As shown in Tab.~\ref{tab:drivinggpt}, our approach achieves significantly stronger results on this benchmark. Due to the absence of released code or full evaluation protocols for GAIA-1 and ADriver-I, we additionally compare against state-of-the-art end-to-end motion planners, where our model demonstrates competitive or superior performance.

\begin{table}[h!]
% \vspace{-4mm}
    \centering
    \caption{Comparison of planning results with DrivingGPT on the NAVSIM test set.}
    % \vspace{-2.5mm}
    \resizebox{\linewidth}{!}{
    \begin{tabular}{l|cc|ccc|c}
        \toprule
        \textbf{Method} & \textbf{NC} $\uparrow$ & \textbf{DAC} $\uparrow$ & \textbf{TTC} $\uparrow$ & \textbf{Comf.} $\uparrow$ & \textbf{EP} $\uparrow$ & \textbf{PDMS} $\uparrow$ \\
        \midrule
        DrivingGPT & \textbf{98.9} & 90.7 & \textbf{94.9} & 95.6 & 79.7 & 82.4 \\
        Ours & 97.9 & \textbf{95.1} & 93.8 & \textbf{99.9} & \textbf{80.4} & \textbf{86.2} \\
        \bottomrule
    \end{tabular}
    }
    \label{tab:drivinggpt}
    \vspace{-2mm}
\end{table}

\paragraph{Comparison with MagicDriveDiT~\cite{gao2024magicdrivedit} and InfinityDrive~\cite{infinitydrive}.}
MagicDriveDiT and InfinityDrive are concurrent works focusing on video generation for autonomous driving. Based on their reported FVD scores on nuScenes (MagicDriveDiT: 94.84, InfinityDrive: 70.06), our method (82.83) exhibits competitive visual generation performance.

We acknowledge that MagicDriveDiT achieves slightly better visual quality, which we attribute to differences in video encoders: they utilize a specialized 3D-VAE, while we adopt a deep-compression autoencoder for better compression and training efficiency. This trade-off may introduce additional visual artifacts, and we plan to improve the DCAE component in future work.

More importantly, as elaborated in Sec.~\ref{subsec:design}, these video diffusion-based methods are designed for scene synthesis without modeling causal dynamics or agent interactions. As a result, they lack support for flexible-length sequence generation and real-time planning, which are crucial in world model settings for decision making and policy learning.

\paragraph{Comparison with Transfusion~\cite{Zhou2024TransfusionPT} and JanusFlow~\cite{Ma2024JanusFlowHA}.}
While these multimodal generative models also combine diffusion and autoregression, their design principles differ significantly from ours. Transfusion and JanusFlow combine \textit{token-wise text autoregression} and diffusion for image understanding and generation. In contrast, our model combine \textit{frame-wise latent autoregression} and diffusion with novel decoupled architecture design to tackle the unique problem of \textit{temporal dynamics and coherence} with video inputs and outputs, which is more challenging.

\section{More Long-term Video Generation Results}
\begin{figure*}[ht]
    \centering
	\includegraphics[width=\linewidth]{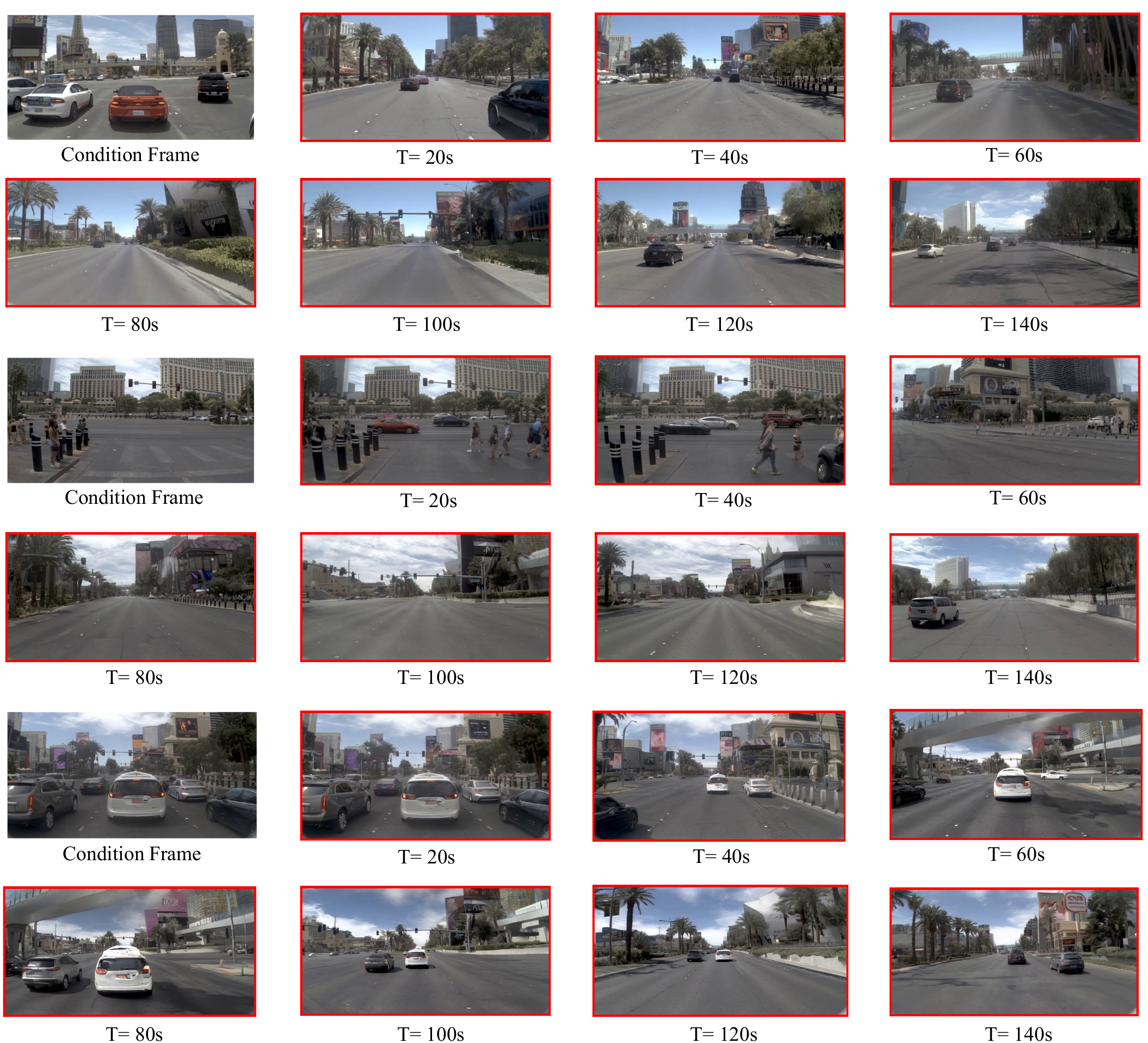}
        % \vspace{-0.3cm}
	\caption{Visualization of Longer Videos. Our world model is capable of generating extended videos (140 seconds) while maintaining high visual quality and detailed vehicles and buildings.}
        \vspace{-0.5cm}
	\label{fig:video}
\end{figure*}

As shown in Fig.~\ref{fig:video}, we present the generation of minute-long ultra-long videos while maintaining high-quality visuals and preserving the integrity and details of surrounding buildings and vehicles. Additionally, our world model continuously generates the next frames with new contents without experiencing context drift.

\label{sec:dit}

\end{document}